\documentclass[runningheads]{llncs}
\usepackage[T1]{fontenc}
\usepackage{graphicx}
\usepackage{amsfonts} 
\usepackage{amsmath}
\usepackage{hyperref}

\begin{document}
\title{Task-guided Domain Gap Reduction for Monocular Depth Prediction in Endoscopy}
\titlerunning{Task-guided Domain Gap Reduction for Endoscopy}
\author{Anita Rau\inst{1,2}\thanks{Project conducted while at University College London.} \and
Binod Bhattarai\inst{1,3}$^\star$\and
Lourdes Agapito\inst{1}\and
Danail Stoyanov\inst{1}}
\authorrunning{A. Rau et al.}
\institute{University College London, UK \and
Stanford University, USA \and
University of Aberdeen, UK}
\maketitle              % typeset the header of the contribution
\begin{abstract}
Colorectal cancer remains one of the deadliest cancers in the world. In recent years computer-aided methods have aimed to enhance cancer screening and improve the quality and availability of colonoscopies by automatizing sub-tasks. One such task is predicting depth from monocular video frames, which can assist endoscopic navigation. As ground truth depth from standard in-vivo colonoscopy remains unobtainable due to hardware constraints, two approaches have aimed to circumvent the need for real training data: supervised methods trained on labeled synthetic data and self-supervised models trained on unlabeled real data. However, self-supervised methods depend on unreliable loss functions that struggle with edges, self-occlusion, and lighting inconsistency. Methods trained on synthetic data can provide accurate depth for synthetic geometries but do not use any geometric supervisory signal from real data and overfit to synthetic anatomies and properties.
This work proposes a novel approach to leverage labeled synthetic and unlabeled real data. While previous domain adaptation methods indiscriminately enforce the distributions of both input data modalities to coincide, we focus on the end task, depth prediction, and translate only essential information between the input domains. Our approach results in more resilient and accurate depth maps of real colonoscopy sequences. The project is available here: \url{https://github.com/anitarau/Domain-Gap-Reduction-Endoscopy}

\end{abstract}
\section{Introduction}
Colorectal Cancer is treatable if detected early, but patient outcome relies on the skill of the performing colonoscopist and complete diagnostic examination of the colon. To improve navigation during colonoscopy and assist endoscopists in ensuring complete examination, computer-assisted mapping and 3D reconstruction could help detect missed surfaces manifesting as holes in the colon map reconstruction \cite{armin2016automated,freedman2020detecting}. Such surgical 3D environment maps could also be used for robotic systems and automation, but despite rapid advances in endoscopic artificial intelligence systems for polyp detection \cite{ahmad2021establishing}, mapping technologies remain challenging to implement robustly. Traditional methods require reliable features to be matched between frames, but colonoscopic images suffer from illumination inconsistency and a lack of texture. 
\begin{figure}[t]
\centering
\includegraphics[width=\linewidth]{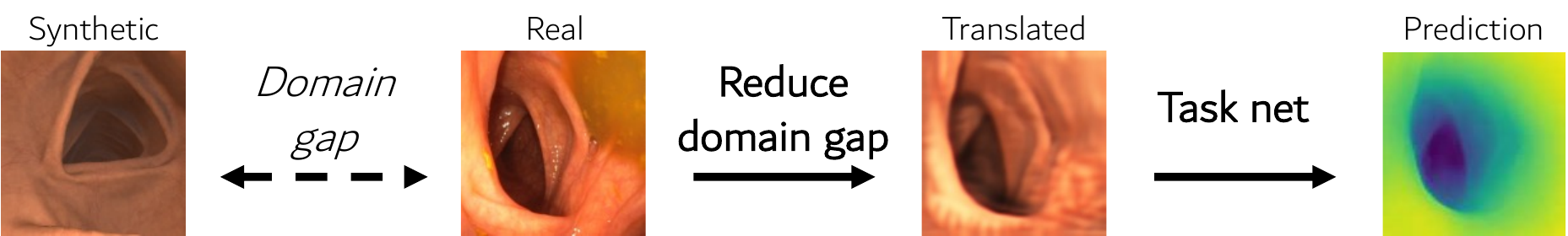} \\
\caption{The proposed network reduces the domain gap between synthetic and real images without fully closing it. We translate only domain- and task-specific information like water which is present in real images but not in synthetic ones.}
\label{fig:teaser}
\end{figure}
A featureless way to obtain a 3D model of the colon is to directly learn frame-wise depth and the relative camera pose between frames. But obtaining ground truth training data for real colonoscopy frames is currently unfeasible, as this would require a depth sensor to be integrated into a standard colonoscope. Instead, self-supervised methods \cite{ozyoruk2021endoslam,shao2022self} do not require any ground truth data and use warping-errors to optimize depth and pose predictions mutually.  While such methods work well on homogeneous surfaces \cite{turan2018unsupervised}, they are challenged by the self-occluding tubular shape of the colon and the view-dependent illumination during colonoscopy. 

\begin{figure}[t]
\centering
\includegraphics[width=\linewidth]{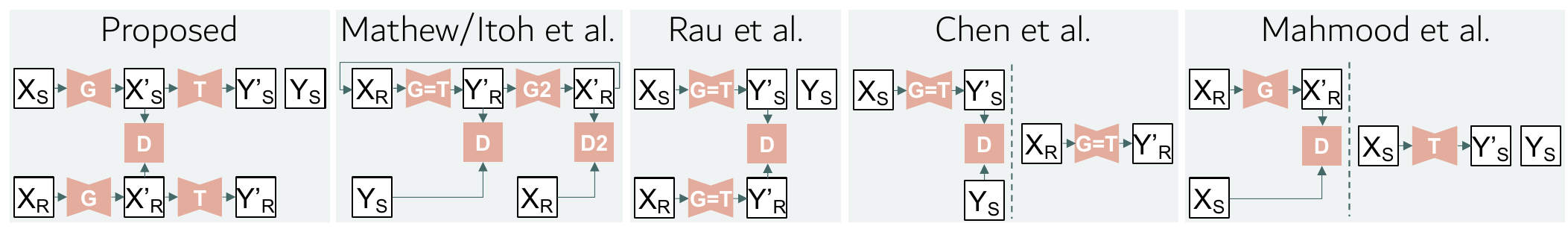}\\
\footnotesize $X_{S/R}$ = Synthetic/Real image, $X'$ = Generated image, $Y_{S/R}$ = Synthetic/Real depth, \\$Y'$ = Generated depth.
$G$ = Generator, $T$ = Task (depth) net, $D$ = Discriminator.\\
\caption{Comparisons of different domain adaptation methods (\cite{rau2019implicit,mahmood2018deep,mathew2020augmenting,itoh2021unsupervised,cheng2021depth}) for depth prediction in colonoscopy. Depiction inspired by \cite{zheng2018t2net}.} 
\label{fig:archicomp}
\end{figure}

An alternative to unlabeled real data is synthetically generated data with ground truth depth. Chen \textit{et al.} propose to first train a network on synthetic data only and in a second, independent step, train the initialized network on real images with self-supervision \cite{cheng2021depth}. However, this approach does not account for the domain shift between real and synthetic images. Other methods have used Generative Adversarial Networks (GANs) to reduce the appearance domain gap, some of which Figure \ref{fig:archicomp} depicts. Mahmood \textit{et al.}~\cite{mahmood2018deep} propose a 
multi-stage pipeline first mapping real examples to the synthetic domain, followed by an independent depth network trained on synthetic data only. 
However, independently training each sub-net might lead to sub-optimal results.
Integrating domain adaptation and depth prediction into a mutual framework, Rau \textit{et al.}~\cite{rau2019implicit} propose to train a single network on real and synthetic images. Mathew \textit{et al.} propose a variation of a cycle GAN that maps virtual images to real images and vice versa \cite{mathew2020augmenting}. One common drawback of these GAN-based methods is the holistic translation from one domain to another without considering domain- and task-specific components. Itoh \textit{et al.} \cite{itoh2021unsupervised} are more deliberate about their choice of translation and decompose information based on a Lambertian-reflection mode; however, this hand-crafted decomposition is not guaranteed to extract and translate the most helpful information. All the translations mentioned above are difficult and distract from the main objective: predicting depth.

Rather than aligning one domain to another, images should reduce the domain gap between real and synthetic images only to the extent that it benefits the end task \cite{pnvr2020sharingan}. Our approach is end-to-end trainable and learns from real and synthetic data accounting for their different geometries by using separate depth losses. As Figure \ref{fig:teaser} shows, the resulting network translates unknown geometric structures like water which is not present in the synthetic dataset. To the best of our knowledge, our method is the first to integrate synthetic
data through a domain-adaptation framework into self-supervised depth estimation in colonoscopy.

\section{Methods}

A standard GAN-based approach to depth prediction can map real and synthetic images to  the real domain \cite{zheng2018t2net}, the synthetic/depth domain \cite{rau2019implicit}, or both \cite{itoh2021unsupervised,mathew2020augmenting}. 
To map an image $X_1\in\mathcal{X}_1$ to a different domain $\mathcal{X}_2$, $X_1$ is passed through a generator $\mathcal{G}$. The output $\mathcal{G}(X_1)$ is then passed through the discriminator $\mathcal{D}$ which compares it to known images from $\mathcal{X}_2$. Minimizing
\begin{align}
\label{Eq:baseline_gan}
        \mathbb{E}_{\mathcal{X}_2}[\text{log}(\mathcal{D}(X_2))] + \mathbb{E}_{\mathcal{X}_1} [\text{log}(1 -\mathcal{D}(\mathcal{G}(X_1)))],
\end{align}
forces $\mathcal{G}$ to learn the distribution of $\mathcal{X}_2$ (\cite{itoh2021unsupervised,mathew2020augmenting,zheng2018t2net,rau2019implicit}).

There are two issues with this approach: (i) these GANs assume that real and synthetic depths come from the same distribution, which is not necessarily true; (ii) the domain adaptation is not guided by the end task. Depth losses for synthetic data are employed but there is no geometric supervision for predicted real depths. The domain adaptation
network thus has no incentive to translate images such that the most accurate
real depths are predicted. Our approach solves both issues.
\subsection{Domain Gap Reduction}
Our method maps as little information as possible to a mutual domain, allowing the network to focus on the end task. This concept was proposed by SharinGAN \cite{pnvr2020sharingan} for depth prediction from calibrated stereo cameras in urban settings. 
%SharinGAN addresses these issues in an urban environment \cite{pnvr2020sharingan}. The GAN translates domain- and task-specific information only and incorporates task losses from both real and synthetic images that drive the task-guided domain gap reduction. The network more accurately predicts depth from unlabeled, real, urban data than previous domain adaptation methods for depth prediction that map images holistically to one of the two domains \cite{zheng2018t2net}. However, SharinGAN requires stereo images, which cannot be acquired with standard colonoscopes. 
\begin{figure*}[t] 
\includegraphics[width=\textwidth]{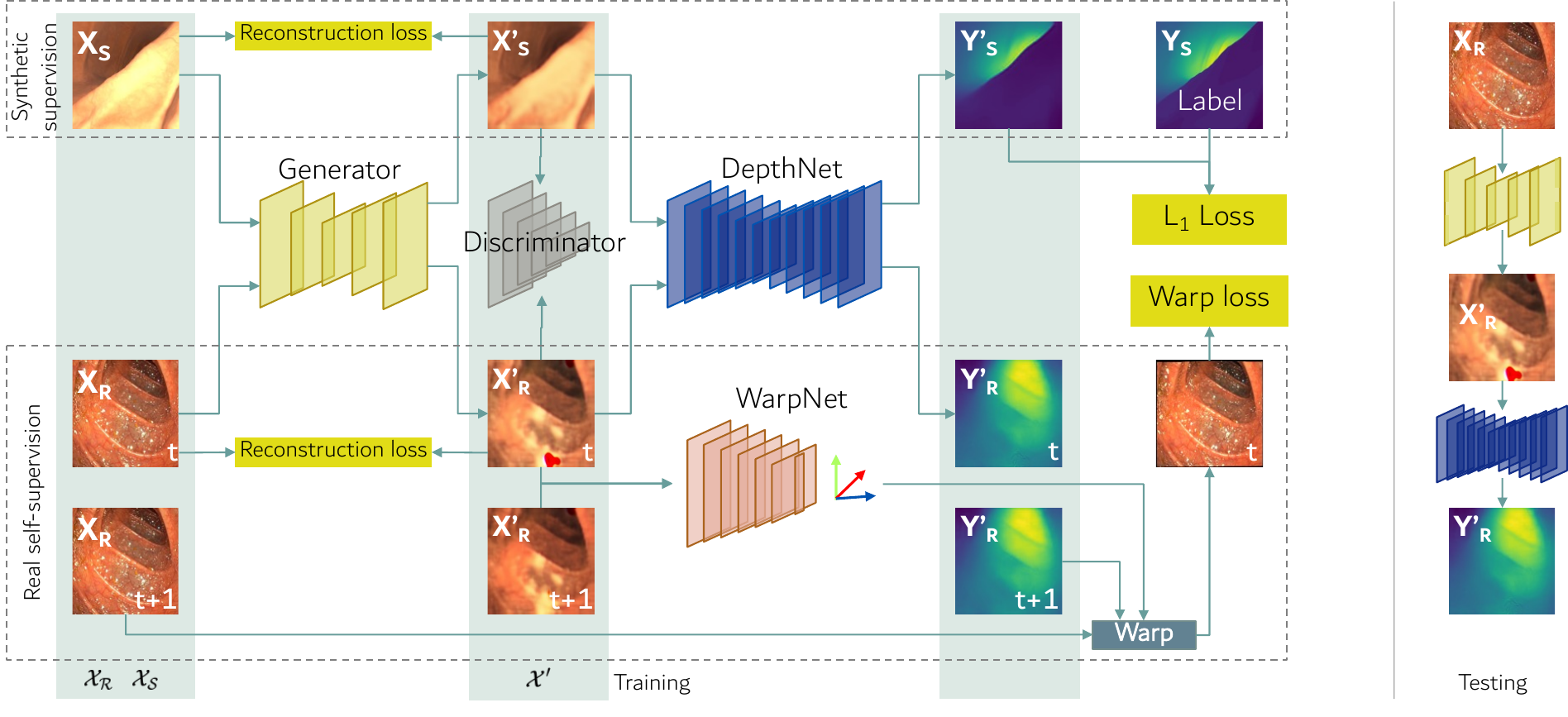}
%\vspace{-2em}
\caption{Overview of our training and testing pipelines. Our network maps real and synthetic images into a new domain $\mathcal{X'}$. These translated images are passed through the DepthNet. Synthetic images are supervised with an $L_1$ error; real images are self-supervised using a warp loss. During testing, inputs are passed through the Generator and DepthNet only.} \label{fig:archi}
\end{figure*}
Figure \ref{fig:archi} shows an overview of our approach. First, a GAN maps synthetic and real images to a mutual and end-task-specific domain. Being in the same domain, the synthetic and real images can now learn depth-specific features from one another. %One could map both domains to the synthetic domain as in \cite{mahmood2018deep}, however, appearing synthetic is not inherently beneficial for the depth prediction task and would skew an aggregated loss functions toward a task that will not benefit the end task. Instead, the generator should only translate those features affecting the depth net while keeping all other information constant.
Let $\mathcal{X_R}$ and $\mathcal{X_S}$ denote the real and synthetic domain and let each image $\in\{\mathcal{X_R},\mathcal{X_S}\}$ consist of domain-agnostic and domain-specific information. 

Domain-agnostic information $I$ is shared between the domains. Such information should encompass the underlying geometry of the colon. We actively avoid adaptation of $I$ as it is unnecessary.
Domain-specific information could be blood vessels that are visible in real images but not in synthetic ones.
Such domain-specific information can be end-task-specific, $\delta_R$ and $\delta_S$, or unspecific, $\delta'_R$ and $\delta'_S$. The end task in our case is depth prediction but other tasks can be adapted to the same concept.
Blood vessels do not encode relevant information about the geometry and are $\delta'_R$; water and shadows, on the other hand, contain information about shape and are $\delta_S$.  The domain gap between  $\delta'_R$ and $\delta'_S$ is negligible, as the depth net will learn to ignore such information. But the domain gap between $\delta_R$ and $\delta_S$ will affect the training of the depth network. If $\delta_R$ and $\delta_S$ are first mapped into a shared domain, real and synthetic data can complement each another.
%Excluding $\delta'_R$ and $\delta'_S$ from the shared domain will ensure that only useful information is translated. 

Let $x$ be a feature of image $X$. We want to learn a mapping $f: \mathcal{X_R} \cup \mathcal{X_S} \rightarrow \mathcal{X'};  x \mapsto f(x)$,
such that $f(x) = x $ if $x \in \{\delta'_S, \delta'_R, \mathcal{I}\}$, and  $f(x) \neq x $ if $x \in \{\delta_S, \delta_R\}$. But how do we learn such a mapping?

Instead of mapping one domain to the other, both domains can be mapped into a mutual one moving the means of the distributions together \cite{arjovsky2017wasserstein}:
\begin{align}
    L_{GAN}= \mathbb{E}_{\mathcal{X_S}}[\mathcal{D}(\mathcal{G}(X_S))]-  \mathbb{E}_{\mathcal{X_R}} [\mathcal{D}(\mathcal{G}(X_R))].
\end{align}
To translate only crucial information in an image while retaining most of it, we use a reconstruction loss that penalizes translation by comparing the generator's input with its output:
\begin{align}
    L_R &=   \|\mathcal{G}(X_S) - X_S\|^2_2 + \|M(\mathcal{G}(X_R) - X_R)\|_1.\label{eq:recon_loss}
\end{align}
We experimentally found the $L1$-loss to lead to more similar reconstructions of small details in the real images and applied a specularity mask $M$ based on the real images' RGB values. 

Now, instead of having to learn how to translate a synthetic image to the real domain, or vice-versa, the network only needs to solve how to translate some information. To encourage that only task-relevant information is translated, we pass the generator's output through a depth net. The depth losses from \textit{both} domains must then be back-propagated as described in the next section.

\subsection{Depth supervision}

As labels for synthetic data exist, synthetic depths are supervised with an $L_1$-loss between the prediction and ground truth:
\begin{align}
    L_S &= \|Y_S' - Y_S\|_1. \label{eq:l1_loss}
\end{align}
But as we miss ground truth for real data, the supervision for the real domain is less straight-forward. SharinGAN proposes to use stereo images for supervision. But in endoscopy we have to  fall back to monocular video. We therefore propose to incorporate self-supervised geometric supervision for real images.
Self-supervised losses help generalize to real anatomies but tend to converge to local minima. Additional synthetic supervision can help guide the optimization of self-supervised models. %We, therefore, incorporate self-supervision for monocular images into our model, giving rise to two depth losses. 

For warping-based self-supervision we pass a second image, $X^{t+1}$, through the same generator and subsequently input both images into a WarpNet, which outputs a 6D pose vector $\mathbf{p}$ allowing us to warp image $X^{t+1}$ to look like image $X^t$. We refer to this warped image as $X^{t+1\rightarrow{t}}$. The warp loss $L_W$ is computed as proposed in \cite{ozyoruk2021endoslam} allowing a direct comparison of both models: 
\begin{align}
    L_W &= 1 * L_{\text{photo}} + 0.5 * L_{\text{geo}} + 0.1 * L_{\text{smooth}}. \label{eq:warp_loss}
\end{align}
It consists of a photometric loss comparing an image to its warped counterpart:
\begin{align}
    L_{\text{photo}} = \sum & \|\mathbf{T}(X^t) - X^{t+1\rightarrow{t}}\|_2, %+ \nonumber \\&0.85 \cdot \frac{1-\text{SSIM}(X^t, X^{t+1\rightarrow{t}})}{2},
\end{align}
where $\mathbf{T}$ is the brightness-aware transformation of $X$ according to \cite{ozyoruk2021endoslam}. Unlike \cite{ozyoruk2021endoslam} we do not incorporate the structural similarity index measure (SSIM) in the warp loss \cite{wang2004image}, making SSIM a fair evaluation measure on the test set. %Note that our WarpNet in Figure \ref{fig:archi} is equivalent to \textit{pose networks} in related works, but as the camera intrinsics of the real images are noisy, we do not claim to predict relative camera poses. 
The geometric consistency loss is based on $Y'^t$ warped to $t+1$, and $Y'^{t+1}$ backwards interpolated to $\tilde{Y}'^{t+1}$ and the smooth loss supports convergence:
\begin{align}
L_{\text{geo}} = \frac{\|Y'^{t\rightarrow{t+1}} - \tilde{Y}'^{t+1}\|_1}{Y'^{t\rightarrow{t+1}}+ \tilde{Y}'^{t+1}}, \quad \text{and } \quad L_{\text{smooth}} = \sum (\exp^{-\nabla X^t} \cdot\nabla Y'^t)^2.
\end{align}

\noindent The final loss is a sum of the GAN-loss, reconstruction loss, and depth losses:
\begin{align}
    L &= \omega_G L_{GAN} + \omega_R L_{R} + 0.5 \cdot (\omega_S L_{S} + \omega_W L_{W}). \label{eq:final_loss}
\end{align}
%Our model is end-to-end trainable and does not require any pre-training or fine-tuning. 
Now that the task losses from {both} domains are back-propagated through the generator, the domain adaptation is guided by the end task and issue (ii) is addressed. Lastly, we observe that our network does not assume that real and synthetic depths are identically distributed (issue i); Figure \ref{fig:archicomp} illustrates that we only input RGB images ($X_S,X_R$) to a mutual discriminator, not depths.
\subsection{Implementation details}
Our DepthNet is the architecture used both in EndoSLAM \cite{ozyoruk2021endoslam} and SharinGAN \cite{pnvr2020sharingan}, allowing a direct comparison of the methods. For further comparability, we use the WarpNet proposed in \cite{ozyoruk2021endoslam}. We use SharinGAN's generator but replace the transposed convolutional layers with interpolation-based upsampling. We replace SharinGAN's discriminator with the lightweight discriminator proposed in Pix2Pix \cite{isola2017image} reducing training time by almost half to 8 hours on one NVIDIA A100-80GB GPU. The loss weights are chosen based on grid search and are: $\omega_G=1, \omega_R=10, \omega_S=100, \omega_W=1$. We train our network on 3,162 image pairs generated from 1,300 real colonoscopy frames of the EndoMapper\footnote{\url{https://www.synapse.org/Synapse:syn26707219/wiki/615178}} dataset \cite{azagra2022endomapper}. All training images were extracted from a single video, as only two videos in the dataset provide camera intrinsics, and one was held out for testing. The synthetic dataset consists of 11,000 frames from the \textit{Unity}-based SimCol\footnote{\url{https://www.ucl.ac.uk/interventional-surgical-sciences/simcol3d-data}} dataset \cite{rau2022bimodal}.
\section{Experiments}

Evaluating a method that bypasses the need for training data is not straightforward, because the absence of test data is inherent to the task. Our evaluation thus first focuses on a qualitative comparison. We then quantitatively compare various reprojection losses for baselines trained in a self-supervised fashion. Lastly, we show that our method generalizes across patients and datasets.
\begin{figure}
\centering
\includegraphics[width=\textwidth]{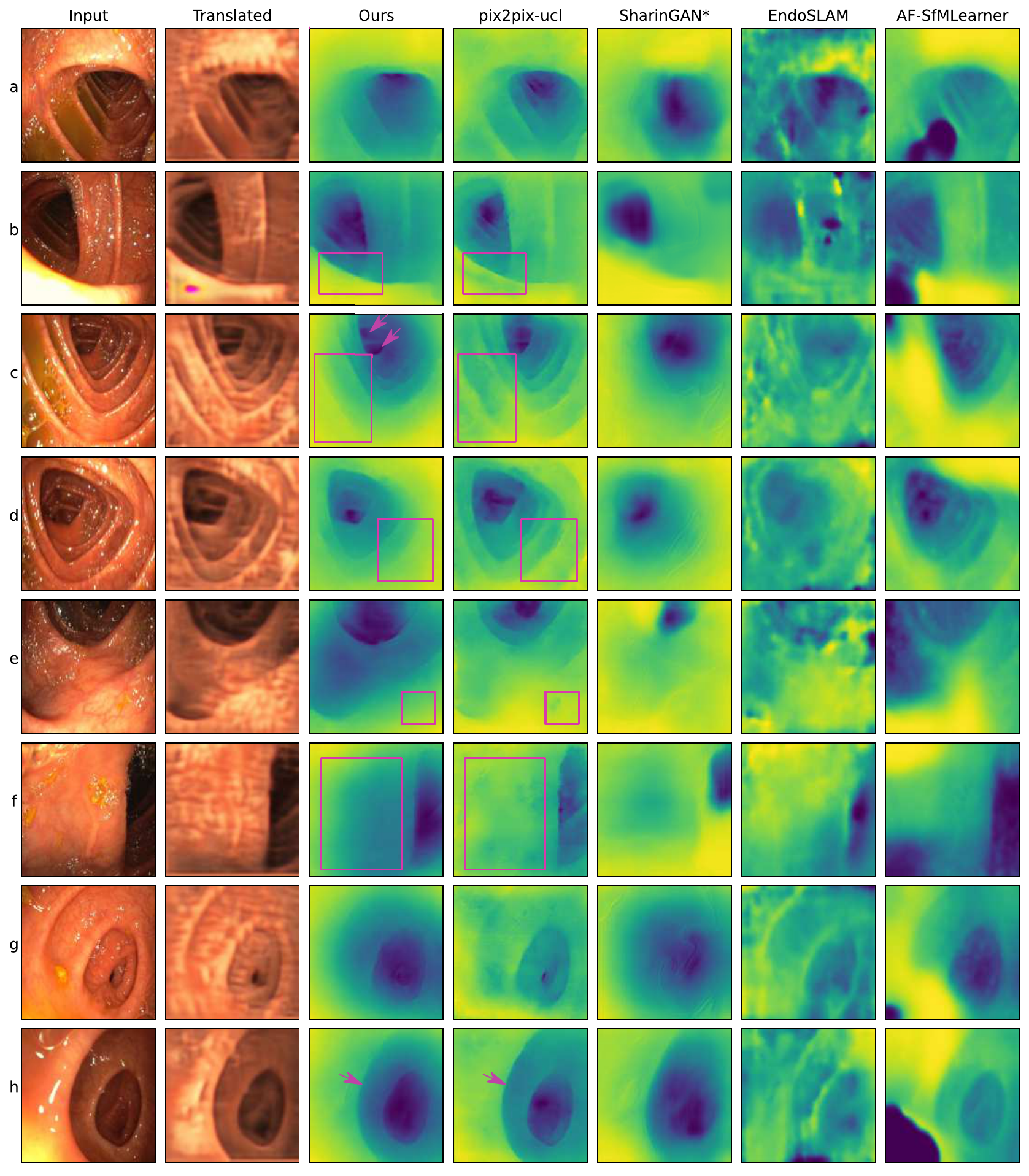} \\
\footnotesize{*SharinGAN without virtual supervision of real images as no stereo data is available.}
\caption{Comparison of different methods on test images. EndoSLAM and the variation of SharinGAN fail to generalize to test data. AF-SfMLearner generalizes more robustly but suffers from large artefacts. We highlight some inconsistencies in pix2pix-ucl  through  magenta boxes. Our method is more resilient to specular highlights, water, and bubbles than the baselines and leads to smoother depth maps where the geometry is even (box in f) while preserving crisp edges (arrow in h) and details (arrows in c).  }
\label{fig:depths}
\end{figure}
\begin{figure*}
\centering
\includegraphics[width=0.45\textwidth]{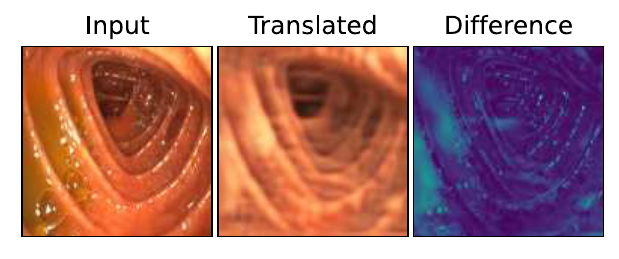}
\includegraphics[width=0.45\textwidth]{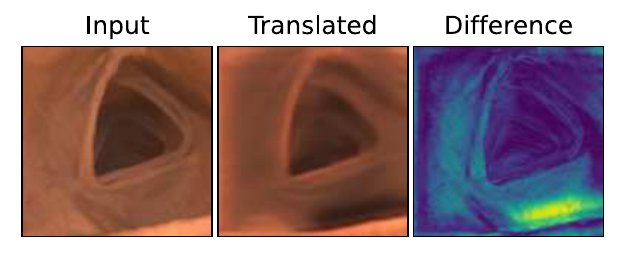}
\caption{Examples of input test images, their translated counterparts, and a difference map, where yellow denotes larger difference. The GAN mainly translates specularities and water in the real image, and shadows in synthetic images (see bottom right corner).}
\label{fig:diff}
\end{figure*}

\noindent \textbf{Qualitative comparison:}
We compare our method to two self-supervised approaches and two domain-adaptation-based algorithms. Our baselines are the self-supervised approaches EndoSLAM \cite{ozyoruk2021endoslam} and AF-SfMLearner \cite{shao2022self} with all parameters set to their default values. The domain-adaptation-based baselines are: (1) a modification of SharinGAN \cite{pnvr2020sharingan}, \textit{SharinGAN*}, in which we omit the virtual supervision of the real images; (2) the extension of Pix2Pix \cite{isola2017image}, referred to as pix2pix-ucl \cite{rau2019implicit}. Figure \ref{fig:depths} depicts results on real test images. These test images are from the same patient but different sections of the colon than the train images.
EndoSLAM fails to generalize to unseen scenes. Although the network converged on the training data, it fails to predict useful depth maps on test images. AF-SfMLeraner predicts largely sensible depth maps but struggles with artifacts like water, stool, and specular highlights. SharinGAN* predicts the overall shape well but fails to preserve details. Pix2pix-ucl, preserves details but the resulting depth maps are patchy and sometimes inaccurate. See, for instance, the highlighted inconsistency in map (e). Further, the gradient of these depth maps is uneven, with only very few pixels assuming high depths (mostly in wrong locations). Our method is the most robust one. It learns from synthetic images, such as SharinGAN or pix2pix-ucl, but incorporating the warping loss helps understand structures that would otherwise be misinterpreted. 

In Figure \ref{fig:diff} we investigate how our GAN works. We plot an image, its translated version, and a difference map for a real and a synthetic example. We can observe that only domain-specific and task-relevant information is translated. In the real image, specularities and water are translated the most (yellow). Specularities encapsulate information about surface normals, while water puddles have specific geometric properties that are not present in the synthetic data. The synthetic example shows that the network hallucinates a strong shadow in the lower right corner, because \textit{Unity's} renderer does not produce entirely realistic shadows.

\noindent \textbf{Quantitative comparison:}
\begin{table}[t]
    \centering
    \caption{Comparison of the different warping-based methods on 1,006 real test images.}

    \begin{tabular}{|l|c|c|c|}
    \hline
    & Photo. loss (Eq. 6) $\downarrow$ & Geo. loss (Eq. 7) $\downarrow$ & SSIM $\uparrow$\\
    \hline
        AF-SfMLearner&  {.096} $\pm$ .081 $\ddagger$&  {.069 }$\pm$ .040 & .686 $\pm$ .134 $\dag$\\
        \hline
        EndoSLAM & {.076} $\pm$ .035 $\dag$& .061 $\pm$ .033 $\dag$ & .641 $\pm$ .104 $\dag$\\
        \hline
        \textbf{Proposed}  &.076 $\pm$ .036 $\dag$& .036 $\pm$ .031 $\dag$& .659 $\pm$ .110\\
        \hline
        
    \end{tabular}\\
    $\dag$) Loss used for training as well. $\ddagger$) Related loss  used for training as well.
    \label{tab:warp}

\end{table}
For methods using warping during training, we can evaluate the warping losses on real test images. These indirectly give us information about the accuracy of the depth \cite{cheng2021depth}. As our network is supervised by the warping loss \textit{and} the synthetic $L_1$-loss, one might assume that our network results in a larger warp error than EndoSLAM, which is trained with warp loss only. However, in Table \ref{tab:warp} we observe that our network results in a comparable photometric loss but a significantly smaller geometric loss. Comparing the SSIM between the methods, we observe that our model produces higher structural similarity than EndoSLAM, although EndoSLAM uses an SSIM-loss during training and our approach does not. A direct comparison of the self-supervised model EndoSLAM and our GAN suggests that synthetic data benefits our shared training approach. AF-SfMLearner produces the highest SSIM, though it was trained with the SSIM-loss while the proposed method was not. This evaluation has limitations. Warping errors evaluate the quality of depth and pose prediction mutually, and the two tasks can compensate each other. We also investigated EndoMapper's provided point clouds as potential pseudo ground truth but found them too noisy and sparse to be useful.

\noindent \textbf{Generalization to new patients:}
\begin{figure}[t]
\centering
\includegraphics[width=\textwidth]{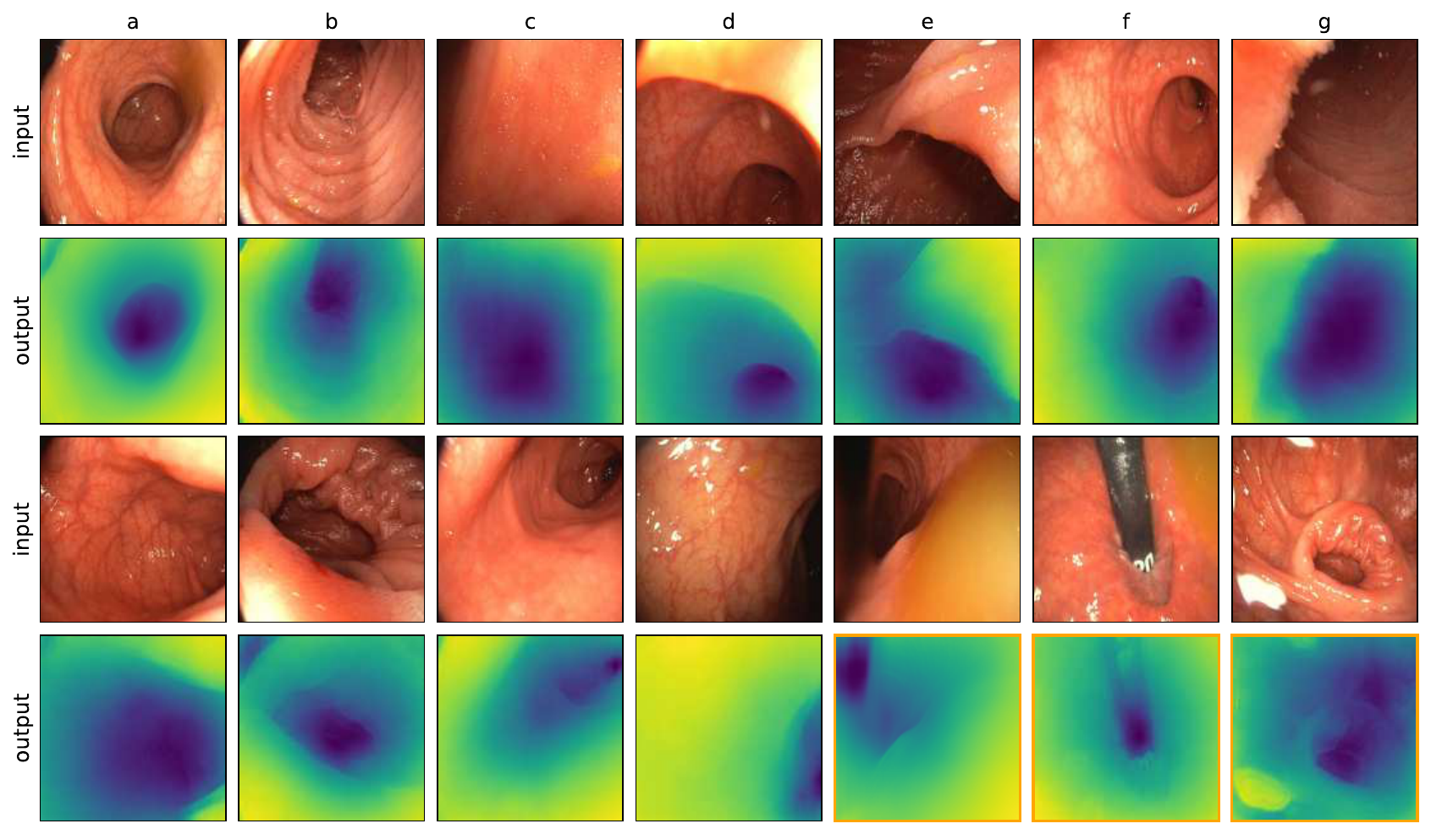} \\
\caption{Generalization to different patients. Our method predicts accurate depths even when trained on only one other patient. Failure cases due to extreme shade, tools, and very large specular highlights, that are not in the training set, are indicated by orange frames.}
\label{fig:generalization}
\end{figure}
The EndoMapper dataset provides COLMAP results for two of the patients. These pseudo-labels are helpful as they provide camera intrinsics and because COLMAP rejects images that are too blurred or too occluded and thus neither useful to extract features nor for our purposes. We found that training on fewer but qualitatively better sequences improves results. We trained our model on one of the two patients with COLMAP labels and evaluated it on the other. Results are shown in Figure \ref{fig:generalization}. We can observe that the model generalizes well to a different patient, even when the colon is filled with water as in image (g, top row) or when geometries are peculiar as in image (b, bottom row). We also show failure cases in the bottom row. In image (e), the model does not generalize to extensive shadow, probably cause by an occluded light source. In image (f), the model falsely locates the retroflexed scope viewing itself in the background. In image (g), the model does not generalize to the extraordinary large specular highlight. However, none of these extreme cases were present in the training data. Nonetheless, the model predicts sharp, accurate, and robust depth maps on most frames of an unseen patient, even when trained on a single anatomy and procedure only.

\begin{figure}[t]
\centering
\includegraphics[width=\textwidth]{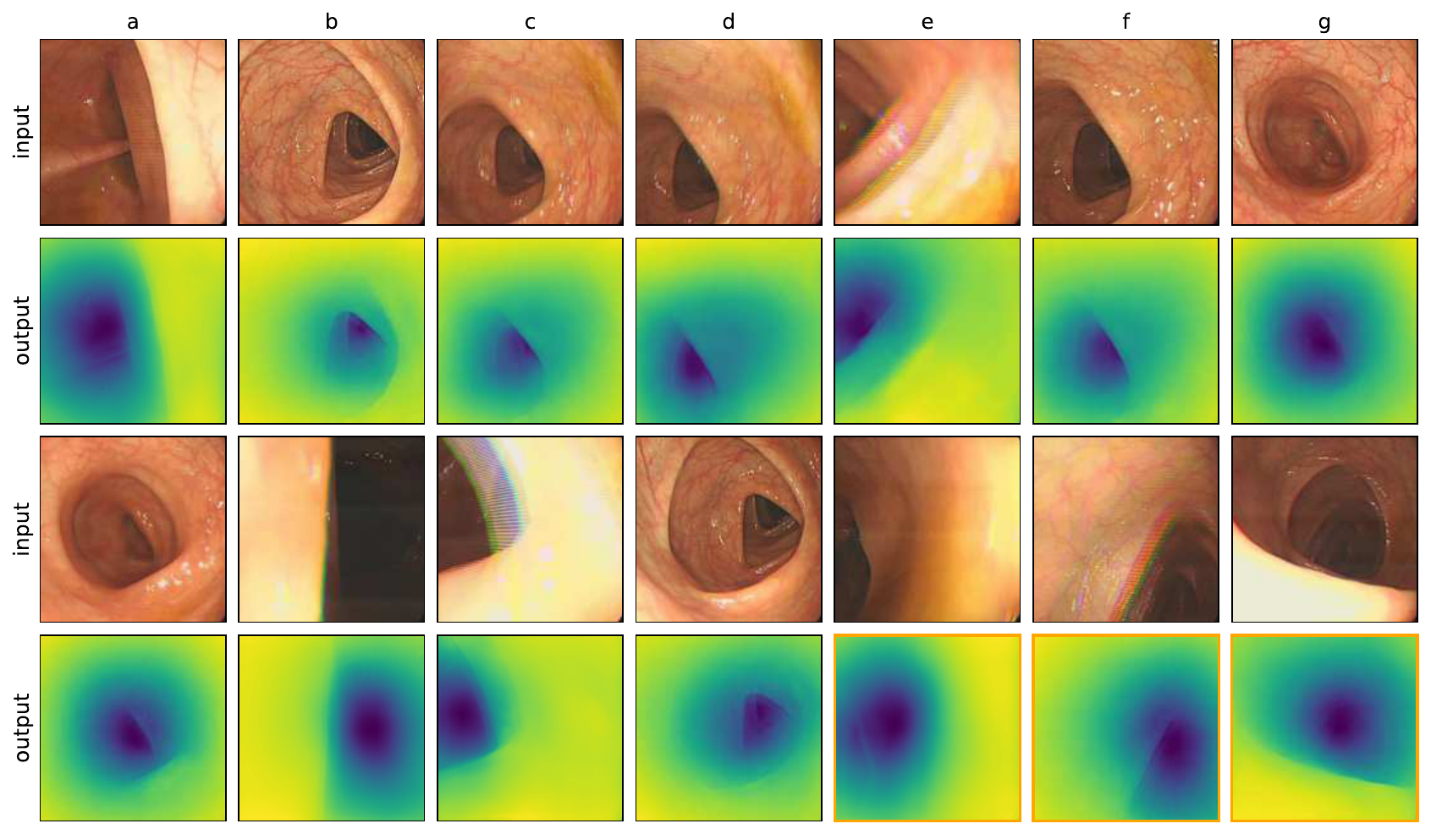} \\
\caption{Depth predictions on the LDPolypVideo dataset. Depicted are results on test images from a different procedure than the training data. Depths are accurate and robust to interlacing artefacts. Failure cases are indicated by orange frames.}
\label{fig:newdata}
\end{figure}

\noindent \textbf{Generalization to different datasets:}
We repeat our experiments on a second publicly available dataset: the LDPolypVideo\footnote{\url{https://github.com/dashishi/LDPolypVideo-Benchmark}} dataset \cite{ma2021ldpolypvideo}, a dataset for polyp detection that conveniently offers polyp-free colonoscopy videos. These videos can be used for our purposes as most frames focus on the lumen rather than the mucosa. As the dataset does not provide camera intrinsics we cannot rule out that the network learns a consistent but skewed geometry. We trained our model on frames from one colonoscopy sequence and applied it to images of a different procedure. We use the same synthetic dataset and hyper-parameters as in the previous experiments. But unlike the EndoMapper dataset, the sequences used for this experiment are only a few minutes long and show only a small section of a colon. Accordingly, the model is only trained on a fraction of the geometries observed in our first experiment. Nonetheless, the model can generalize to a different patient predicting accurate and sharp depth maps and is highly robust to interlacing artefacts as illustrated in Figure \ref{fig:newdata}.

\section{Conclusions}

%In this work, we propose a method for task-guided domain gap reduction for depth prediction in colonoscopy. We show that our model only translates information in an input image that contributes to the depth prediction task. Using synthetic and real images in such a mutual framework leads to more robust and accurate depth predictions than standard self-supervised approaches. Synthetic training data is more easily obtainable than (unlabeled) real data and can be leveraged for the depth prediction task in real scenarios. Nonetheless, our approach shows that the correct integration of synthetic and real data is crucial. Compared to pure domain-adaptation systems, self-supervision helps the network adapt to geometric properties of real data that are not present in synthetic data. 

%Our experiments show that our approach leads to geometrically consistent depths, outperforming previous domain translation methods and self-supervised models. We demonstrate that our results are more consistent with the smooth surfaces of the colon, more robust to unseen geometries, like water and air bubbles, and still preserve details and edges. In the future, other tasks could benefit from task-guided domain gap reduction, as long as task losses are available for both domains.

Learning-based depth prediction has seen significant advances in recent years but requires labels for training, which are not available for colonoscopy. This work addresses the question how unlabeled real data and cheap, labeled synthetic data can be used in a mutual framework without overfitting to the geometry of the synthetic data. At the core of this work is the idea that domain adaptation is a challenging task that should only be addressed to the extent that it benefits the end task. Rather than indiscriminately translating entire images from one domain to another, and accounting only for appearance domain gaps, we propose task-guided domain gap reduction.

Our experiments show that our model learns to translate only task- and domain-specific information in real and synthetic input images. The network learns that water and air bubbles are specific to real data and that rendered shadows in synthetic data differ from real data. Accounting for these task-specific differences leads to geometrically consistent depth maps, outperforming previous domain translation and self-supervised models. We demonstrate that our results are more consistent with the smooth surfaces of the colon, more robust to unseen geometries, and still preserve details and edges. In the future, other tasks could benefit from task-guided domain gap reduction.

\subsubsection{Acknowledgements} This work was supported by the Wellcome/EPSRC Centre for Interventional and Surgical Sciences (WEISS) [203145Z/16/Z]; Engineering and Physical Sciences Research Council (EPSRC) [EP/P027938/1, EP/ R004080/1, EP/P012841/1]; The Royal Academy of Engineering Chair in Emerging Technologies scheme; and the EndoMapper project by Horizon 2020 FET (GA 863146). All datasets used in this work are publicly available and linked in this manuscript. The code for this project is publicly available on Github.
For the purpose of open access, the author has applied a CC BY public copyright license to any author-accepted manuscript version arising from this submission.

\bibliographystyle{splncs04}%
\bibliography{sn-bibliography}

\begin{thebibliography}{10}
\providecommand{\url}[1]{\texttt{#1}}
\providecommand{\urlprefix}{URL }
\providecommand{\doi}[1]{https://doi.org/#1}

\bibitem{ahmad2021establishing}
Ahmad, O.F., et~al.: Establishing key research questions for the implementation of artificial intelligence in colonoscopy: a modified delphi method. Endoscopy  \textbf{53}(09),  893--901 (2021)

\bibitem{arjovsky2017wasserstein}
Arjovsky, M., Chintala, S., Bottou, L.: Wasserstein generative adversarial networks. In: ICML. pp. 214--223. PMLR (2017)

\bibitem{armin2016automated}
Armin, M.A., et~al.: Automated visibility map of the internal colon surface from colonoscopy video. IJCARS  \textbf{11}(9),  1599--1610 (2016)

\bibitem{azagra2022endomapper}
Azagra, P., et~al.: Endomapper dataset of complete calibrated endoscopy procedures. arXiv preprint arXiv:2204.14240  (2022)

\bibitem{cheng2021depth}
Cheng, K., et~al.: Depth estimation for colonoscopy images with self-supervised learning from videos. In: MICCAI. pp. 119--128 (2021)

\bibitem{freedman2020detecting}
Freedman, D., et~al.: Detecting deficient coverage in colonoscopies. IEEE TMI  \textbf{39}(11),  3451--3462 (2020)

\bibitem{isola2017image}
Isola, P., Zhu, J.Y., Zhou, T., Efros, A.A.: Image-to-image translation with conditional adversarial networks. In: CVPR. pp. 1125--1134 (2017)

\bibitem{itoh2021unsupervised}
Itoh, H., et~al.: Unsupervised colonoscopic depth estimation by domain translations with a lambertian-reflection keeping auxiliary task. IJCARS  \textbf{16}(6),  989--1001 (2021)

\bibitem{ma2021ldpolypvideo}
Ma, Y., et~al.: Ldpolypvideo benchmark: A large-scale colonoscopy video dataset of diverse polyps. In: MICCAI. pp. 387--396 (2021)

\bibitem{mahmood2018deep}
Mahmood, F., Durr, N.J.: Deep learning and conditional random fields-based depth estimation and topographical reconstruction from conventional endoscopy. Medical image analysis  \textbf{48},  230--243 (2018)

\bibitem{mathew2020augmenting}
Mathew, S., Nadeem, S., Kumari, S., Kaufman, A.: Augmenting colonoscopy using extended and directional cyclegan for lossy image translation. In: Proceedings of CVPR. pp. 4696--4705 (2020)

\bibitem{ozyoruk2021endoslam}
Ozyoruk, K.B., et~al.: Endoslam dataset and an unsupervised monocular visual odometry and depth estimation approach for endoscopic videos. Medical image analysis  \textbf{71},  102058 (2021)

\bibitem{pnvr2020sharingan}
PNVR, K., Zhou, H., Jacobs, D.: Sharingan: Combining synthetic and real data for unsupervised geometry estimation. In: Proceedings of CVPR. pp. 13974--13983 (2020)

\bibitem{rau2022bimodal}
Rau, A., Bhattarai, B., Agapito, L., Stoyanov, D.: Bimodal camera pose prediction for endoscopy. IEEE Transactions on Medical Robotics and Bionics pp.~1--1 (2023). \doi{10.1109/TMRB.2023.3320267}

\bibitem{rau2019implicit}
Rau, A., et~al.: Implicit domain adaptation with conditional generative adversarial networks for depth prediction in endoscopy. IJCARS  \textbf{14}(7),  1167--1176 (2019)

\bibitem{shao2022self}
Shao, S., et~al.: Self-supervised monocular depth and ego-motion estimation in endoscopy: Appearance flow to the rescue. Medical image analysis  \textbf{77},  102338 (2022)

\bibitem{turan2018unsupervised}
Turan, M., et~al.: Unsupervised odometry and depth learning for endoscopic capsule robots. In: IROS. pp. 1801--1807 (2018)

\bibitem{wang2004image}
Wang, Z., Bovik, A.C., Sheikh, H.R., Simoncelli, E.P.: Image quality assessment: from error visibility to structural similarity. IEEE transactions on image processing  \textbf{13}(4),  600--612 (2004)

\bibitem{zheng2018t2net}
Zheng, C., Cham, T.J., Cai, J.: T2net: Synthetic-to-realistic translation for solving single-image depth estimation tasks. In: Proceedings of the European Conference on Computer Vision. pp. 767--783 (2018)

\end{thebibliography}
\end{document}